\documentclass[10pt,twocolumn,letterpaper]{article}

\usepackage[pagenumbers]{cvpr} 

\usepackage{graphicx}
\usepackage{amsmath}
\usepackage{amssymb}
\usepackage{booktabs}
\usepackage{times}
\usepackage{epsfig}
\usepackage{enumitem}
\usepackage{breqn}
\usepackage{mathtools}
\newcommand{\OURNAME}{VGFlow\xspace}

\usepackage[pagebackref,breaklinks,colorlinks]{hyperref}

\usepackage[capitalize]{cleveref}
\crefname{section}{Sec.}{Secs.}
\Crefname{section}{Section}{Sections}
\Crefname{table}{Table}{Tables}
\crefname{table}{Tab.}{Tabs.}

\def\cvprPaperID{4509} 
\def\confName{CVPR}
\def\confYear{2022}

\begin{document}

\title{\OURNAME: Visibility guided Flow Network for Human Reposing }

\author{Rishabh Jain\thanks{rishabhj@adobe.com}\\
MDSR Adobe\\
\and
Krishna Kumar Singh\\
Adobe Research\\
\and
Mayur Hemani\\
MDSR Adobe\\
\and
Jingwan Lu\\
Adobe Research\\
\and
Mausoom Sarkar\\
MDSR Adobe\\
\and
Duygu Ceylan\\
Adobe Research\\
\and
Balaji Krishnamurthy\\
MDSR Adobe\\
}

\maketitle


\begin{abstract}
The task of human reposing involves generating a realistic image of a person standing in an arbitrary conceivable pose. There are multiple difficulties in generating perceptually accurate images, and existing methods suffer from limitations in preserving texture, maintaining pattern coherence, respecting cloth boundaries, handling occlusions, manipulating skin generation, etc. These difficulties are further exacerbated by the fact that the possible space of pose orientation for humans is large and variable, the nature of clothing items is highly non-rigid, and the diversity in body shape differs largely among the population. To alleviate these difficulties and synthesize perceptually accurate images, we propose \OURNAME. Our model uses a visibility-guided flow module to disentangle the flow into visible and invisible parts of the target for simultaneous texture preservation and style manipulation. Furthermore, to tackle distinct body shapes and avoid network artifacts, we also incorporate a self-supervised patch-wise "realness" loss to improve the output. \OURNAME achieves state-of-the-art results as observed qualitatively and quantitatively on different image quality metrics (SSIM, LPIPS, FID). \textbf{All results would be made public}
\end{abstract}
\section{Introduction}
\label{sec:intro}
People are frequently featured in creative content like display advertisements and films. As a result, the ability to easily edit various aspects of humans in digital visual media is critical for rapidly producing such content. Changing the pose of humans in images, for example, enables several applications, such as automatically generating movies of people in action and e-commerce merchandising. This paper presents a new deep-learning-based framework for reposing humans guided by a target pose, resulting in high-quality and realistic output.

\begin{figure}[h!]
\begin{center}
  \includegraphics[width=\linewidth]{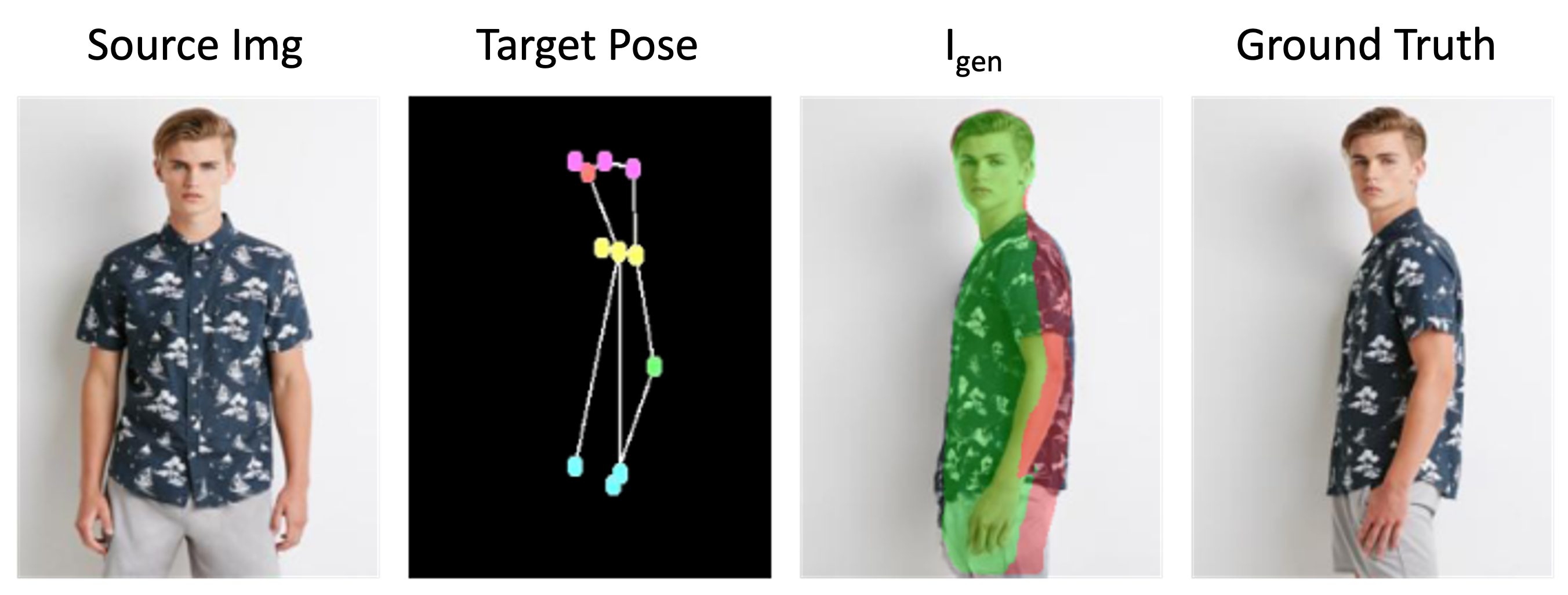}
\end{center}
\caption{Human reposing involves changing the orientation of a source image to a desired target pose. To get accurate results, we learn to preserve the region visible (green) in the source image and transfer the appropriate style to the invisible region (red)} 
\label{fig:generator}
\end{figure}

Recent approaches for human-image reposing based on deep-learning neural networks, such as \cite{casd, nted, spgnet}, require a person image, their current pose, represented as a sequence of key-points or a 2D projection of a 3D body-pose map,  and the target pose represented similarly. These methods fail to reproduce accurate clothing patterns, textures, or realistic reposed human images. This mainly happens when either the target pose differs significantly from the current (source) pose, there are heavy bodily occlusions, or the garments are to be warped in a non-rigid manner to the target pose. Many of these failures can be attributed to the inability of these networks to discern regions of the source image that would be visible in the target pose from those that would be invisible. This is an important signal to determine which output pixels must be reproduced from the input directly and which must be predicted from the context. We present \OURNAME, a framework for human image reposing that employs a novel visibility-aware detail extraction mechanism to effectively use the visibility input for preserving details present in the input image.

\OURNAME consists of two stages - encoding the changes in appearance and pose of the source image required to achieve the new pose and decoding the encoded input to the re-posed human image. The encoding stage includes a pose-based warping module that takes the source image and the source and target pose key-points as input and predicts two 2D displacement fields. One corresponds to the visible region of the source image in the target pose, and the other to the invisible areas. It also predicts a visibility mask indicating both visible and invisible regions in the source image, as they should appear in the target pose. The displacement fields, known as \textit{appearance flows}, are used to sample pixels from the source image to produce two warped images. These warped images and the visibility masks are then encoded into the appearance features, a multi-scale feature pyramid. The encoding stage also tries to capture the relationship between the source and target poses by encoding their respective key-points together. The encoded pose key-points are translated into an image during the decoding stage, with the appearance features modulating the translation at each scale. This appearance-modulated pose to image decoding provides the final reposed output, which is then subjected to multiple perceptual and reconstruction losses during training. 

The vast majority of existing methods\cite{gfla, dior, nted, casd} are trained using paired source and target images. However, in terms of output realism, we observe various artifacts and a lack of generalization in these methods to unpaired inputs, especially when the source image differs significantly in body shape or size\cite{ma2021fda}. To that end, \OURNAME is trained with a self-supervised patch-wise adversarial loss on unpaired images alongside the pairwise supervised loss to ensure a high level of realism in the final output.
In summary, this paper proposes a new human reposing network \OURNAME, based on:
\begin{itemize}[nosep]
\item A novel visibility-aware appearance flow prediction module to disentangle visible and invisible regions of the person image in the target pose.
\item An image decoder employing multi-scale texture modulated pose encoding.
\item And, a patch-wise adversarial objective to improve the realism of the produced images leading to fewer output artifacts.
\end{itemize}

Our method achieves state-of-the-art on image quality metrics for the human reposing task. We present extensive qualitative and quantitative analysis with previous baselines, as well as ablation studies. Next, we discuss work related to the proposed method.
\section{Related work}


\paragraph{Human Reposing} In recent years, several works have tried to generate a person in the desired target pose~\cite{casd, ren2021flow, adgan, song2019unsupervised, albahar2021pose, zhang2020cross, liu2019liquid}. One of the initial work was $PG^2$~\cite{pg2}, which concatenated the source image with the target pose to generate the reposed output. Their work produced erroneous results due to the misalignment between the source and target image. Follow-up works tried to mitigate the problem by using a deformable skip connection~\cite{siarohin2018deformable} or progressive attention blocks\cite{zhu2019progressive} to achieve better alignment. However, modeling complex poses and bodily occlusions were still challenging. Recently, there have been some attention-based approaches~\cite{casd, nted} proposed to learn the correspondence between the source and target pose. Although these attention\cite{nted} and style distribution-based\cite{casd} methods have shown impressive results, they need improvement for handling complex transformations. Apart from being computationally expensive, they do not preserve spatial smoothness and are inefficient in modeling fine-grained texture. In contrast, our approach is a flow-based method that can naturally handle complex and large transformations by using flow to warp the person in the source image to the target pose while preserving geometric integrity.

\paragraph{Flow-based methods} Flow estimation aids in learning the correspondence between the content of two images or video frames and has been used in a variety of computer vision tasks such as optical flow estimations~\cite{teed2020raft, xu2022gmflow}, 3D scene flow~\cite{li2021neural}, video-to-video translation~\cite{chen2019mocycle}, video inpainting~\cite{li2022towards}, virtual-try-on~\cite{zflow, han2019clothflow}, object tracking~\cite{fan2021optical} etc. Flow-based methods are also heavily explored for the human reposing task\cite{albahar2021pose, gfla, dior, spgnet} in which pixel-level flow estimates help to warp the texture details from the source image to the target pose. Still, as they don't incorporate visibility cues in their architecture, the network often relies more on in-painting rather than preserving the source image content. DIF\cite{li2019dense} utilized a visibility mask to refine their flow estimation by splitting the appearance features into visible and invisible regions and applying a convolution on top. However, we show that visible and invisible information contain crucial complementary details and should be treated separately in the network pipeline. Most current networks are trained with paired poses, in which the target image is of the same person as the source image. The inference is usually performed for a different human, with the target pose derived from someone with a different body shape and size. This results in unusual distortions in the output. FDA-GAN\cite{ma2021fda} proposed a pose normalization network in which they added random noise to the target pose using SMPL\cite{loper2015smpl} fitting to make reposing robust to noise. However, adding random noise would not lead to the imitation of real human pose distributions. We tackle this problem by introducing a self-supervised adversarial loss using unpaired poses during training.
\section{Methodology}

Our human reposing network requires three inputs, source image, source pose keypoints, and target pose keypoints. The task is to use the appearance and pose information from inputs to generate an image in the target pose. We subdivide this task into two key sub-parts. A detail extraction task, conditioned on the target pose, followed by a generation task where the extracted details are used to create a realistic image.
Our detail extraction task is performed by a flow module $FlowVis$, which warps and aligns the texture and body shape information from the source image with the target pose. The outputs of $FlowVis$ are passed through our generator module, where texture and latent vectors from different resolutions are merged with the pose encoding using 2D style modulation. We also fine-tune the network using self supervised patch-wise realness loss to remove the generator artifacts. More details can be found in subsequent sections. 

%
\begin{figure*}
\begin{center}
  \includegraphics[width=\linewidth]{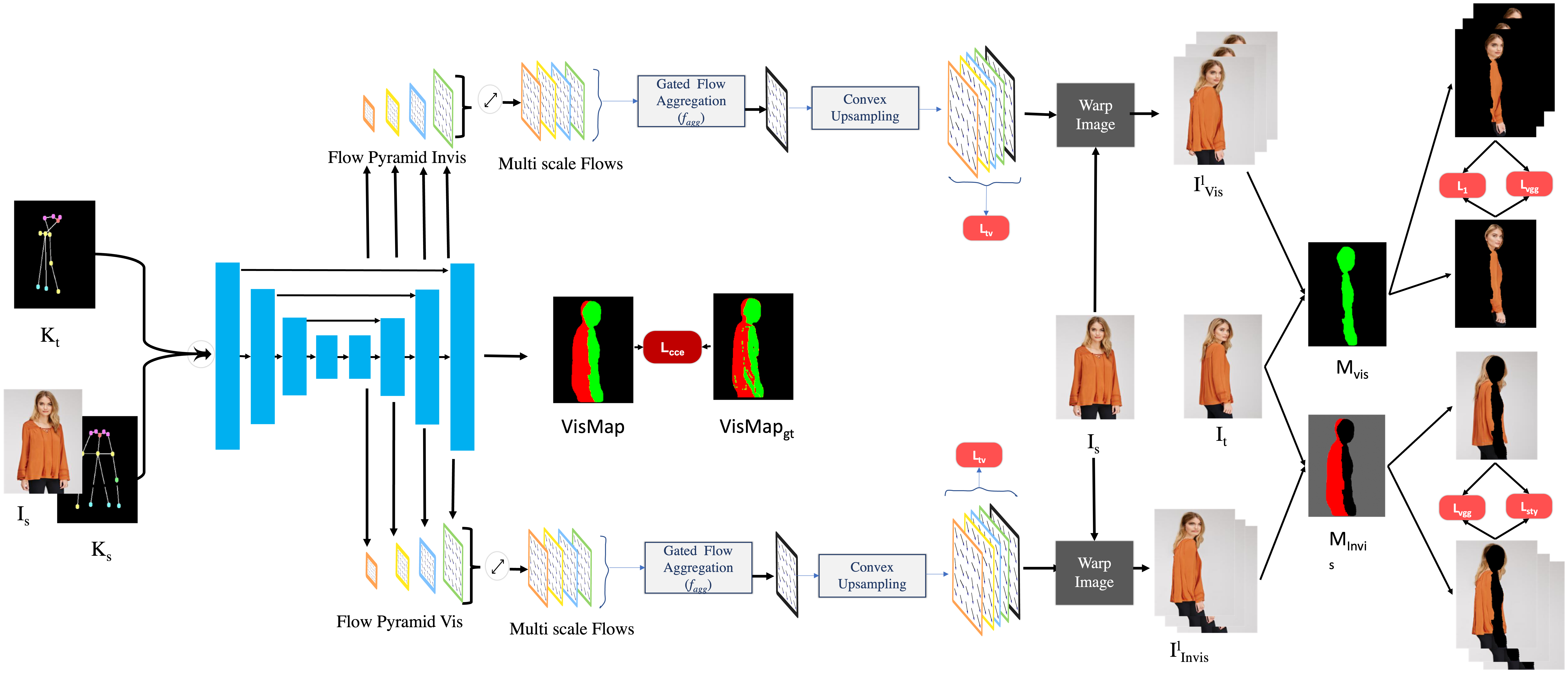}
\end{center}
\caption{Our \textit{FlowVis} module takes in concatenated $I_s, K_s\ \& \ K_t$ as input. We predict two flow pyramids at multiple scales for different visibility regions using a Unet architecture. Subsequently, these flow pyramids are combined using \textit{Gated Aggregation}\cite{zflow} and upsampled via \textit{Convex upsampling}\cite{raft}. Losses $L_1, L_{vgg}$  are imposed on the visible areas and   $L_{vgg}, L_{sty}$ are imposed on the invisible areas.} 
\label{fig:flow}
\end{figure*}

\subsection{Flow \& Visibility module} CNNs are well suited for image-to-image transformation tasks where there is a per-pixel mapping between input and output. An absence of such pixel-to-pixel mapping requires incorporation of other methods to aid learning. Therefore, many human reposing networks \cite{gfla,spgnet,dior,albahar2021pose} usually employ some warping to deform source images to align them with output pose to get a better spatial correspondence between input and output. Warping an image requires a per pixel displacement flow field of size $2\times H \times W$, where $H$ and $W$ are the height and width of the image respectively. This flow field can be learnable or obtained using a 3d model of humans such as SMPL\cite{loper2015smpl} or UV maps\cite{densepose}. The flow generated by existing works are not able to preserve the intricate geometry and patterns present on real clothes. These inadequacies get highlighted in the presence of complex poses and occlusions in the source image. To alleviate these issues, we propose a novel visibility aware flow module.

Our $FlowVis$ Module (Fig \ref{fig:flow}) takes in the source image $I_s$, source pose keypoints $K_s$, and target pose keypoints $K_t$ as inputs and generates a visibility map $VisMap$, two flow field pyramids $f_{v}^l, f_{i}^l$(for visible and invisible regions respectively) and two warped image pyramids $I_{v}^l$, $I_{i}^l$ using the generated flow fields. The $VisMap$, $f_{v}^l$ and $f_{i}^l$ are generated by a Unet\cite{UNet} like architecture FlowPredictor(FP). $VisMap$ segments the target image $I_t$ into visible and invisible regions in the source image (Fig \ref{fig:flow}). The visible region(green in $VisMap$ ) corresponds to an area that is visible in $I_s$, and the invisible part (red) is the area that is occluded in $I_s$. The two separate per-pixel displacement flow-field pyramids $f_{v}^l$  and $f_{i}^l$ are predicted at different resolutions $l$. These flows are used to warp the source image to align with the target pose and generate the target's visible $I_{v}^l$ and invisible $I_{i}^l$ regions. The insight for predicting two flow fields comes from the observation that prediction for both the visible $I_{v}^l$ and invisible $I_{i}^l$ target regions may require pixels from the same location in the source. Therefore we need two flow fields to mitigate this issue because a single flow field can only map a source pixel to either one of the two regions. Flows from multiple resolutions are combined using the Gated aggregation\cite{zflow} (GA) technique which filters flow values from different radial neighborhoods to generate a composite flow. This allows the network to look at multiple scales sequentially and refine them at each step. 
To construct the flow at the final $256\times256$ level, we upsample the flow from the previous layer using convex upsampling. Convex upsampling\cite{raft} is a generalization of bilinear upsampling where the upsampling weights for a pixel are learnable and conditioned on the neighborhood. Convex upsampling aids in the preservation of fine-grained details and sharpens the warped output. Moreover, employing a single decoder to generate both flow pyramids helps preserve consistency and coherence between visible and invisible warped images. The module is summarised in the following equations.
\begin{small}
\begin{equation}
\label{eq:singleViewReposing}
\begin{split}
f_{v}^l, f_{i}^l, VisMap \leftarrow FP(I_{s},  K_{s}, K_{t}) \\
f_{v}^{agg}, f_{i}^{agg} \leftarrow GA(f_{v}^l, f_{i}^l) \\
f_{v}^{o}, f_{i}^{o} \leftarrow ConvexUpsample(f_{v}^{agg}, f_{i}^{agg}) \\
I_{v}^l, I_{i}^l \leftarrow Warp(I_{s}, f_{v}^{l}), Warp(I_{s}, f_{i}^{l})
\end{split}
\end{equation}
\end{small}

\vspace{-4mm}
\paragraph{Losses} We train the flow module seperately before passing the output to the generator. $FlowVis$ is trained to generate $I_{v}$ and $I_{i}$ by minimizing the losses on the visible and invisible regions of the human model respectively. The $VisMap$ can be broken down to visible area mask $m_{v}$ and invisible area mask $m_{i}$ by comparing the per-pixel class. For the visible region, we can find an exact correspondence between predicted and the target image and hence we utilize L1 and perceptual loss $L_{vgg}$ \cite{VGGCNN} on the masked visible part. The loss on the visible area minimizes texture distortion and loss in detail(L1). For the invisible region, there is no exact correspondence between $I_{i}$ and $I_{t}$ but we can still optimize using the target image style. For example, in-case a person needs to be reposed from a front pose to back pose then the entire body will be invisible in $I_{s}$. However, there is a very strong style similarity that we can leverage for reposing. Hence, we use perceptual \cite{VGGCNN} and style loss $L_{sty}$\cite{gatys} to capture the resemblance for these regions on the masked invisible regions. We also minimize tv norm\cite{wedel2009improved} on the flow pyramids to ensure spatial smoothness of flow and the losses are computed for the entire flow pyramid. $Vismap$ is optimized by imposing cateogrical cross entropy loss, $L_{cce}$. The ground truth for the visibility map $Vismap_{gt}$, was obtained by fitting densepose\cite{densepose} on $I_{s}$, $I_{t}$ and then matching the acquired $UV$ coordinates to generate the visible and invisible mask. We further use teacher forcing technique\cite{pratama2021scalable} for training $FlowVis$, in which the ground truth VisMap is used with 50\% probability for the warping losses.
The losses guiding the flow module can be summarized as follows. Here $\odot$ indicates per pixel multiplication.
\begin{small}
\begin{equation}
\label{eq:lossFunction}
\begin{split}
m_{v}, &m_{i} \leftarrow VisMap \\
L_{wrp} =  \sum_{l}^{}L_{vis}&(I^{l}_{v}, m_{v}, f_l) +  L_{invis}(I^{l}_{i}, m_{i}, f_l) \\
 + & L_{cce}(VisMap, VisMap_{gt})
\end{split}
\end{equation}
\end{small}

where,
\begin{small}
\begin{align}
\begin{split}
L_{vis}(I,m,f) &= \beta_{1} L_{vgg}(I\odot m, I_{t}\odot m)\  + \\ &\enskip  \beta_{2} \|I\odot m, I_{t}\odot m\|_1 +  \beta_{3}L_{tv}(f)\\
L_{invis}(I,m,f) &= \beta_{1} L_{vgg}(I\odot m, I_{t}\odot m)  \\ &\enskip +  \beta_{4} L_{sty}(I\odot m, I_{t}\odot m) +  \beta_{3}L_{tv}(f) 
\end{split}
\end{align}
\end{small}

\subsection{Generator module}
The generator module (Fig \ref{fig:generator}) takes as input the source and target poses and the output of $FlowVis$ module. It is important to point out that only the final level of the transformed image pyramid $I_{v}^l,I_{i}^l$, i.e. the level at resolution $256\times256$ referred now as $I_{v},I_{i},$  is used in our generator.

\vspace{-4mm}
\paragraph{Pose encoder} Majority of the previous networks\cite{casd, dior, gfla} encoded the pose information only as a function of 18 channel target keypoints. However, single view reposing is fundamentally an ill defined task. The network has to hallucinate the body and clothing region whenever it is invisible in the source image. Hence, it should be able to distinguish between the portions of the target image for which it can obtain corresponding texture from the source image and those for which it must inpaint the design of the respective clothing item. Therefore, to model the correlation between the source($K_s$) and target poses($K_t$), we pass both source and target keypoints to a Resnet architecture $PoseEnc$ to obtain a $16\times16$ resolution pose feature volume. 

\begin{equation}
e_{p} =  PoseEnc(K_{s}, K_{t})
\end{equation}

\begin{figure}
\begin{center}
  \includegraphics[width=\linewidth]{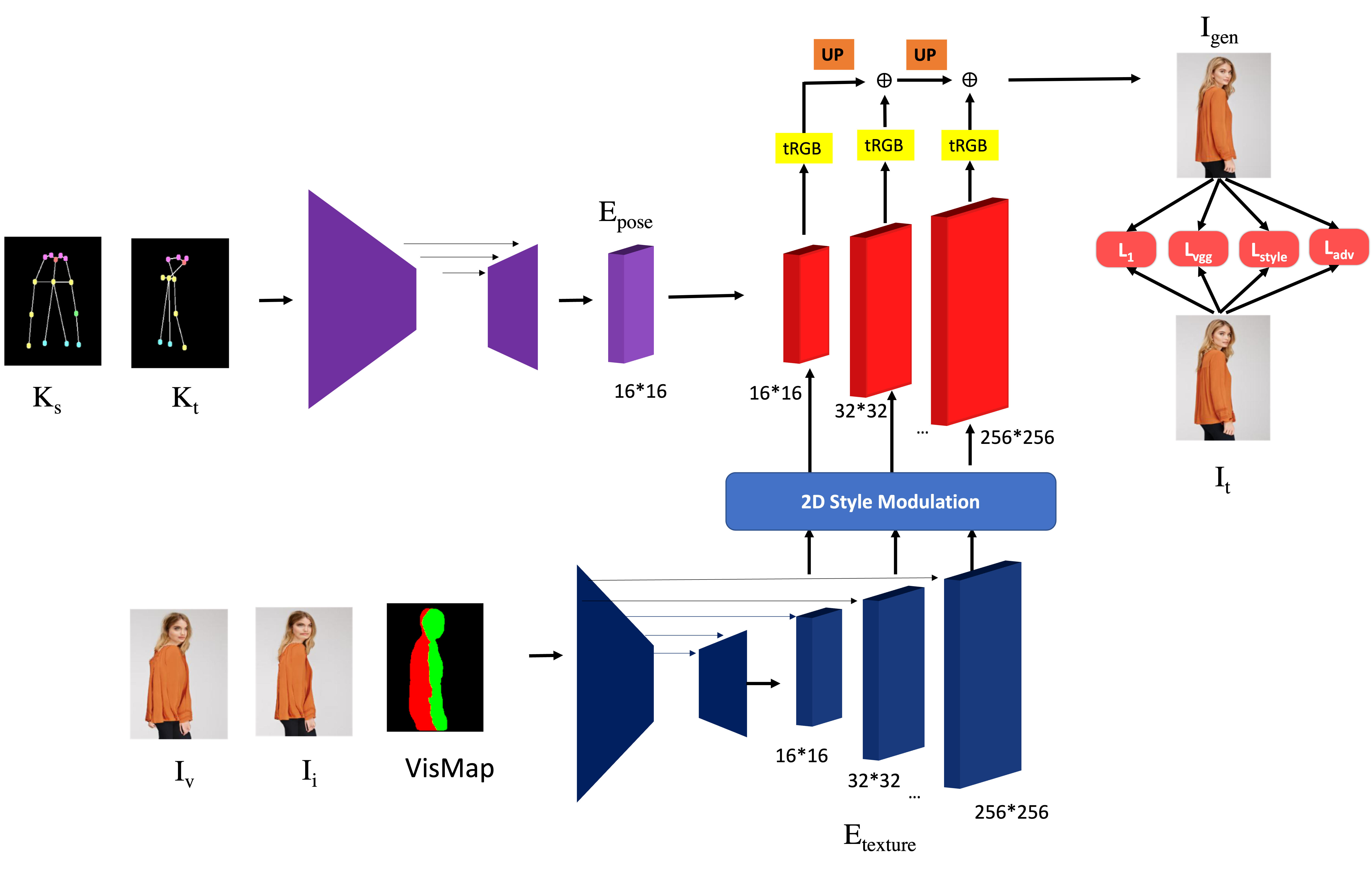}
\end{center}
\caption{Our \textit{Generator} module consumes the FlowVis outputs to generate the final reposed output. It utilizes 2D style modulation\cite{albahar2021pose} to inject Multi-scale Appearance features into the pose encoding for the generation process} 
\label{fig:generator}
\end{figure}

\vspace{-4mm}
\paragraph{Texture injection \& Image Generation}
The texture encoder takes in $I_{v}, I_{i}$ and $Vismap$  as input and uses a ResNet architecture similar to pose encoder to obtain texture encodings at different hierarchical scales. The low resolution layers are useful for capturing the semantics of the clothing items, identity of the person and the style of individual garments while the high resolution layers are useful for encapsulating the fine grained details in the source image.  We also add skip connections in our texture encoder which helps in combining low and high resolution features and capture different levels of semantics.

The image decoder module takes in the pose encoding $e_{p}$ as input and up samples them to higher resolutions. The texture is injected into these pose encodings at different scales by using 2D style modulation\cite{albahar2021pose}.
 After the modulation, features are normalized such that they have zero mean and unit standard deviation. Similar to~\cite{albahar2021pose} which is based on StyleGAN2 architecture, RGB images are predicted at multiple resolutions and sequentially lower resolution image is added to next higher resolution image after upsampling it to obtain the final output image (Fig \ref{fig:generator}).  As the network has to fill in the invisible region ($I_i$) by generating new content similar to neighbourhood pixels at inference, we perform an auxiliary task of inpainting 20\% of the training time, similar to \cite{dior}. Here, a random mask is applied on the target image and given as input to the model. The generator is then asked to output the complete target image. This teaches our network to complete any missing information of warped images in a visually convincing manner.  \OURNAME achieves SOTA in the single view human reposing task. 


\begin{small}
\begin{align}
\begin{split}
e_{tex}^{l} = TexEnc(I_{vis}, I_{Invis}, VisMap) \\
I_{gen} = ResNetDec(e_{p}, 2DStyleMod(e_{tex}^{l}))  
\end{split}
\end{align}
\end{small}

\begin{figure}
\begin{center}
  \includegraphics[width=\linewidth]{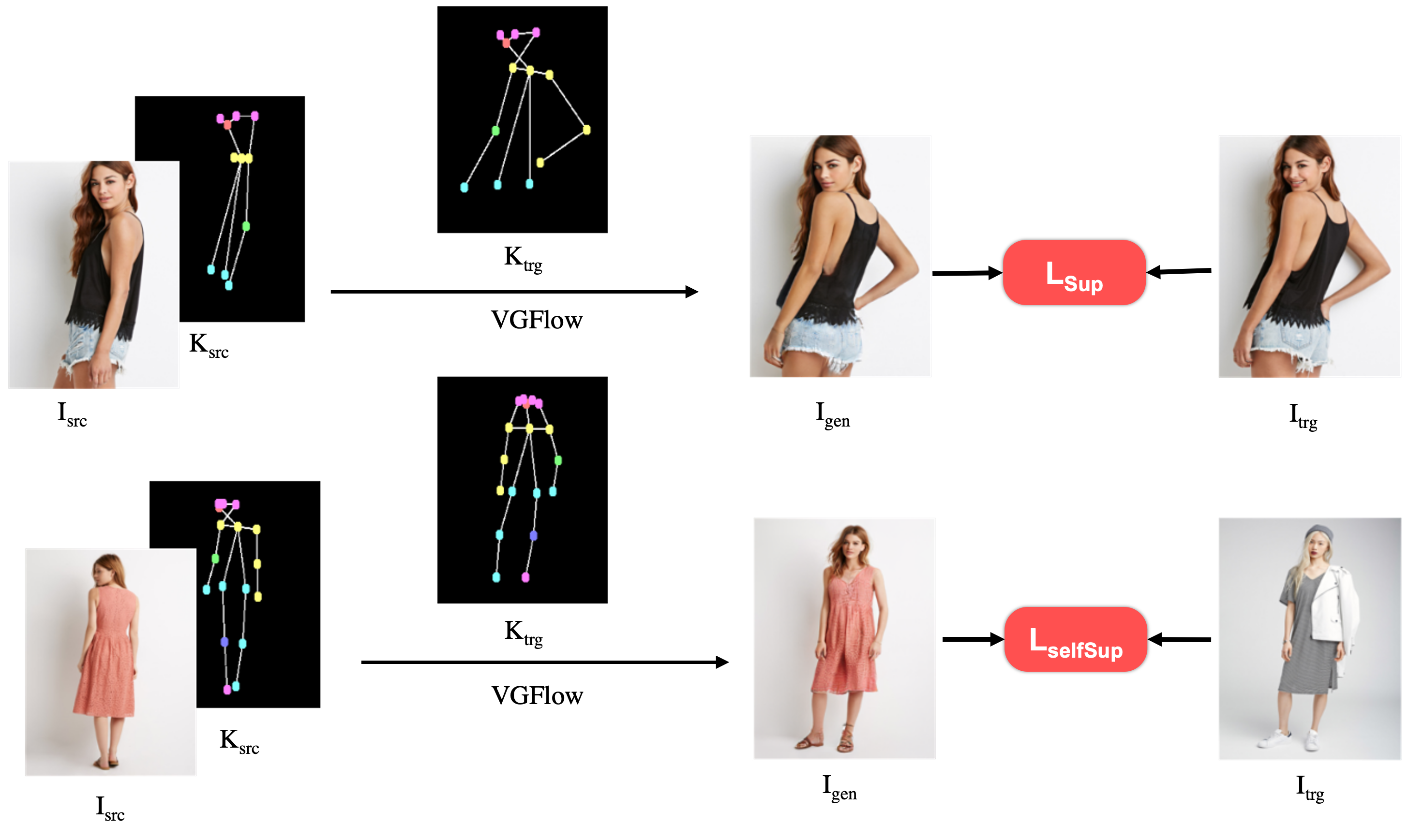}
\end{center}
\caption{Addition of patch-wise Self Supervised loss($L_{SS}$) helps in enhancing the image quality and increases \textit{realness}} 
\label{fig:selfsupLoss}
\end{figure}

\vspace{-4mm}
\paragraph{Losses} We enforce L1, VGG, Style and LSGAN\cite{mao2017least} losses between $I_{gen}$ and $I_{t}$. L1 loss helps to preserve the identity, body pose and cloth texture with pixel level correspondence. Vgg and Style loss are useful in conserving high level semantic information of garments present on source human and bringing the generated image perceptually closer to target. For the LSGAN loss, we pass target pose($K_{t}$) along with generated image($I_{gen}$) to the discriminator for better pose alignment. Adverserial loss assist in removing small artifacts and making the image more sharper. We utilize LSGAN loss as it has been shown to produce better output than traditional GAN loss\cite{mao2017least} and is more stable during training. Overall, the loss function can be defined as:

\begin{small}
\begin{align}
\begin{split}
&L_{sup} =\alpha_{rec} \|I_{gen},I_{t}\|_1  + \alpha_{per} L_{vgg}(I_{gen},I_{t}) \\
&\enskip + \alpha_{sty} L_{sty}(I_{gen},I_{t}) + \alpha_{adv} L_{LSGAN}(I_{gen},I_{t},K_{t}) 
\end{split}
\end{align}
\end{small}
The $\alpha$'s are the weight for the different losses. $L_{sup}$ refers to the supervised loss used when the input and output are of same person in the same attire.
\vspace{-4mm}
\paragraph{Self supervised \textit{realness} Loss} Even though the network is able to produce perceptually convincing results with the above supervised training, there are still bleeding artifacts present that occur when there is a complicated target pose or occluded body regions. There is also a discontinuity present between the clothing and skin interface at some places. Moreover, during training, the target pose is taken from paired image where the input and output are for the same person wearing the same apparel. This would introduce a bias in the network as it would not be robust to alterations in body types(e.g. fat, thin, tall) between $K_s, K_t$\cite{ma2021fda}. During inference this could deteriorate results when, for example, the body shape of the person from which the target pose is extracted vary significantly from the source image human body shape. To alleviate these issues, we fine tune our network with an additional patch wise adversarial loss \cite{patchgan} whose task is to identify if a particular patch is real or not. Therefore during fine-tuning, we choose unpaired images with 50\% probability(Sec \ref{sec:implementationDetails}) from a single batch. Only an adversarial loss is applied on the unpaired images and $L_{sup}$ loss is present on the paired images(Fig \ref{fig:selfsupLoss}). 

\begin{small}
\begin{align}
\begin{split}
&L_{SS} = L_{PatchGAN}(I_{gen},I_{t})
\end{split}
\end{align}
\end{small}
All these losses i.e. $L_{wrp}, L_{sup}$ and $L_{SS}$ are used to  finetune the networks in an end-to-end fashion.

\section{Experiments}
\label{sec:exper}

\paragraph{Dataset} We perform experiments on the In-shop clothes benchmark of Deepfashion\cite{deepfashion} dataset. The dataset consists of 52,712 pairs of high resolution clean images with ~200,000 cross-pose/scale pairs. The images are divided into 48,674 training and 4038 testing images and resized to a resolution of $256\times256$. The keypoints of all images are extracted using OpenPose\cite{openpose} framework. We utilize the standard 101,966 training and 8,570 testing pairs, following previous works\cite{albahar2021pose, dior, nted}. Each pair consists of source and target image of the same person standing in distinct pose. It is also worth noting that the identity of training and testing pairs are separate from one another in the split. 

\vspace{-4mm}
\paragraph{Evaluation metrics} We compute structural similarity index(SSIM), Learned Perceptual Image Patch Similarity (LPIPS) and Fretchet Inception Distance(FID) for comparison with previous baselines. SSIM quantifies image degradation by employing luminance, contrast and and structure present in the image\cite{setiadi2021psnr}. LPIPS\cite{zhang2018unreasonable} compute patchwise similarity using features from a deep learning model. It has been shown to correlate well with human perception of image similarity. FID \cite{heusel2017gans} works by comparing the 2-wasserstein distance between the InceptionNet statistics of the generated and ground truth dataset. This provides a good metric for estimating the \textit{realness} of our generated results. 

\label{sec:implementationDetails}
\vspace{-4mm}
\paragraph{Implementation details} All the experiments were carried out using pytorch framework on 8 A100 gpus. We first train our reposing network using only supervised losses for 30 epochs with Adam\cite{kingma2014adam} optimizer, batch size of 32 and learning rate of 1e-4. Afterwards, we fine tune the model using self supervised loss. For the self supervised training of our model, we randomly choose a target sample or pose image from the complete training dataset during training. We also choose the patch size of $16\times16$ for the discriminator which provides a good tradeoff between high level and low level feature preservation. We found that imposing the self supervised loss works better when we impose it intra-batch. So, supervised and self supervised losses were propagated together in the same mini-batch and we used loss masking to stop gradients for the respective branches. In contrast, the inpainitng auxilliary task was carried out among different mini-batches. The intra-batch strategy provides an anchoring to the network with supervised losses in a single backward pass as we don't have a pairwise supervision for a random target pose. Without intra-batch training, the network starts hallucinating details which are not consistent with the source image but looks plausible. The finetuning was carried out with a batch size of 32 and a reduced learning rate of 5e-5. We highlight additional advantages of using self supervised learning technique in sec \ref{sec:ablations}.
\section{Results}

We compare our method with several strong previous works\cite{gfla, dior, spgnet, albahar2021pose, nted, casd}. Among these, Gfla\cite{gfla}, Spgnet\cite{spgnet}, Dior\cite{dior} and PWS\cite{albahar2021pose} leverage flow-based warping in their network to align source image with the target pose. The flow obtained by them is used to warp the features obtained from the source image and move them to the $k_{t}$ orientation. PWS\cite{albahar2021pose} takes additional information in the form of UV maps of the target image as input. Note that the densepose UV maps of the target image contain a lot more information than simple keypoints. PWS\cite{albahar2021pose} exploits the UV map and symmetry to directly calculate the flow and inpaint the occluded part using a coordinate completion model. However, in the case of unpaired images at inference, obtaining the ground truth UV maps is a difficult task, and the body shape problem is worsened further. Moreover, UV map contains information about the per-pixel mapping of the body regions between $I_{s}$ \& $I_{t}$. Such extensive information won't be accurate if we use a UV map from an unpaired pose and hence, keypoints-based methods are not directly comparable with UV-based methods. We include qualitative results from \cite{albahar2021pose} for completeness and to highlight the issues in UV based warping in Sec.\ref{sec:ablations}. In contrast to flow-based techniques,  NTED\cite{nted} and CASD\cite{casd} utilize semantic neural textures and cross-attention-based style distribution respectively to diffuse the texture of the source image onto the target pose. However, disentangling the texture and reconstructing it in a different orientation is a challenging task without warping assistance, especially for intricate patterns and text. We further corroborate this phenomenon in our qualitative analysis.

\begin{figure*}
\begin{center}
  \includegraphics[width=0.95\linewidth]{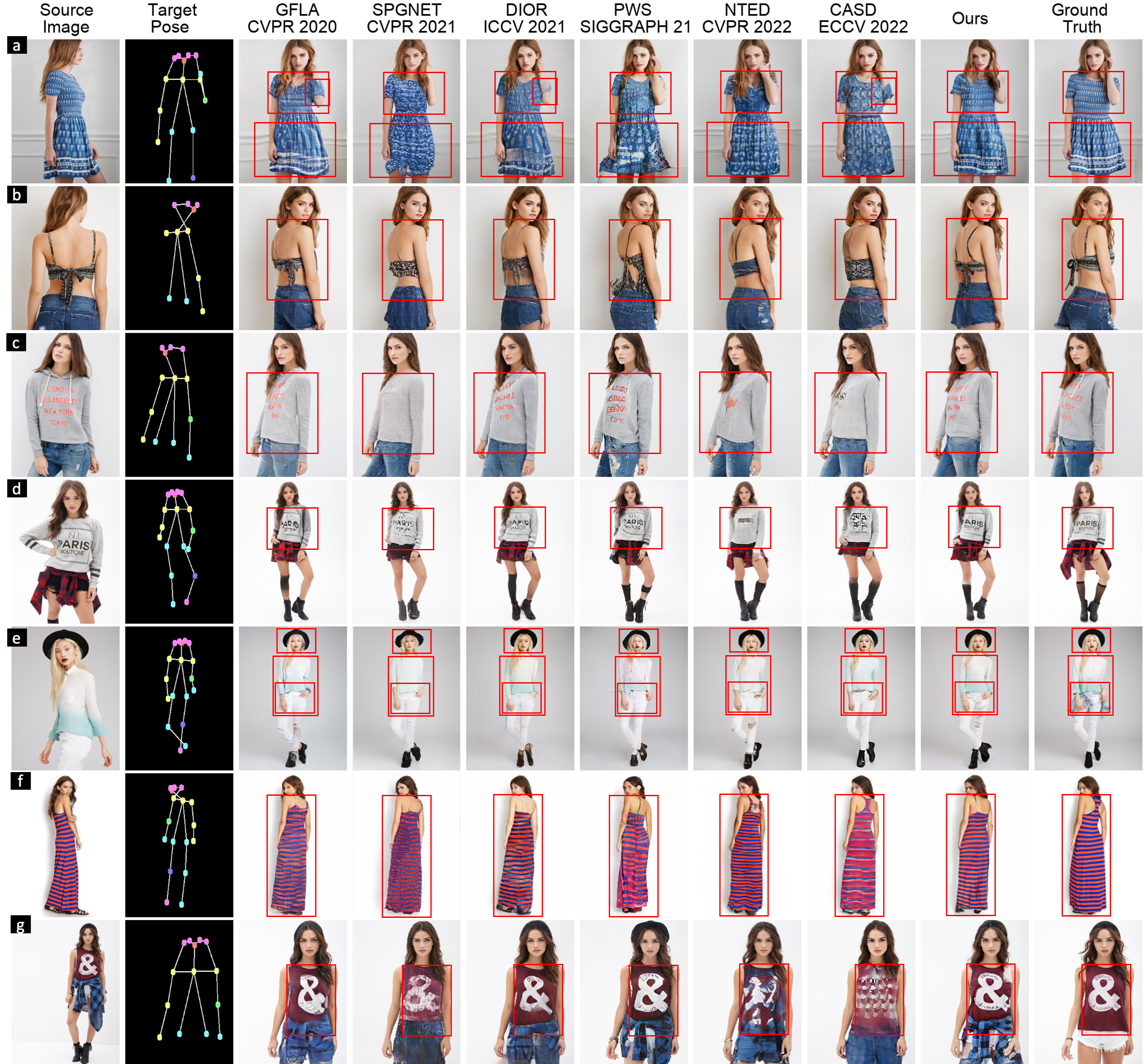}
\end{center}
\caption{In this figure, we underscore improvements along different qualitative aspects over previous works. We emphasize enhancement in preserving pattern \& segmentation (a), handling complex poses (b),  design consistency (c), conserving text readability (d), reducing bleeding, skin generation \& identity reproduction (e), maintaining geometric integrity (f) and pattern coherence (g). (Best viewed in zoom)}
\label{fig:qualitativeResults}
\end{figure*}

\begin{table}[h!]
\begin{center}
\begin{tabular}{|c|c|c|c|}
\hline
Method    & SSIM $\uparrow$  & FID $\downarrow$  & LPIPS $\downarrow$  \\
\hline
Intr-Flow\cite{li2019dense} & - & 16.31 & 0.213 \\
GFLA\cite{gfla} & 0.713 & 10.57 & 0.234 \\
ADGAN\cite{adgan} & 0.672 & 14.45 & 0.228 \\
SPGNet\cite{spgnet} & 0.677 & 12.24 & 0.210 \\
Dior\cite{dior} & 0.725 & 13.10 & 0.229 \\
CASD\cite{casd} & 0.724 & 11.37 & 0.193 \\
Ours & \textbf{0.726} & \textbf{9.29} & \textbf{0.185} \\
\hline
\end{tabular}
\end{center}
\caption{Our network outperforms all the previous baselines for quantitative image metrics at $256\times256$ resolution}
\label{tab:quant-sota}
\end{table}

\vspace{-4mm}
\paragraph{Quantitative comparison for human reposing} We compare our method with previous works which have generated images with $256\times 256$ resolution as their output size. As can be seen from Table \ref{tab:quant-sota}, \OURNAME performs significantly better in SSIM, LPIPS and FID compared to other baselines. The improvement in LPIPS and SSIM metric, which compare images pairwise, can be attributed to better modelling of the visible regions of the model. On the other hand, FID, which measures the statistical distance between the latent space distribution of real and generated images, becomes better due to our superior style transfer for the invisible region.

\vspace{-4mm}
\paragraph{Qualitative comparison for human reposing} We highlight improvements in human reposing from several qualitative dimensions in Fig \ref{fig:qualitativeResults}. In (a), \OURNAME generates images with accurate pattern reproduction and seamless intersection of top and bottom designs. Other works were either not able to maintain coherence during generation or produced unnatural artifacts in their final outputs.  In (b), our network was able to reproduce the back ribbon at the correct target location along with the strap. Whereas, PWS\cite{albahar2021pose} is not able to model the ribbon as the UV map is a dense representation of only the human body and does not capture the details of loose clothes properly. (c) and (d) highlight texture preservation while reposing where the text maintains its relative orientation during warping. The words in the original pose are only intelligible for \OURNAME. In (e), the blue color is bleached onto the white shirt, and face reproduction is not accurate for multiple other reposed outputs. (f) shows the faithful geometric construction of parallel lines and (g) emphasizes pattern coherence while global deformation of texture.  We see that only the warping-based methods were able to preserve the rough texture in (g) while NTED\cite{nted} and CASD\cite{casd} completely dismantled the structural integrity of the``\&" symbol on the t-shirt. Additional results can be found in supplementary material.
\section{Ablation and Analysis}
\label{sec:ablations}

\begin{figure}[t]
\begin{center}
  \includegraphics[width=\linewidth]{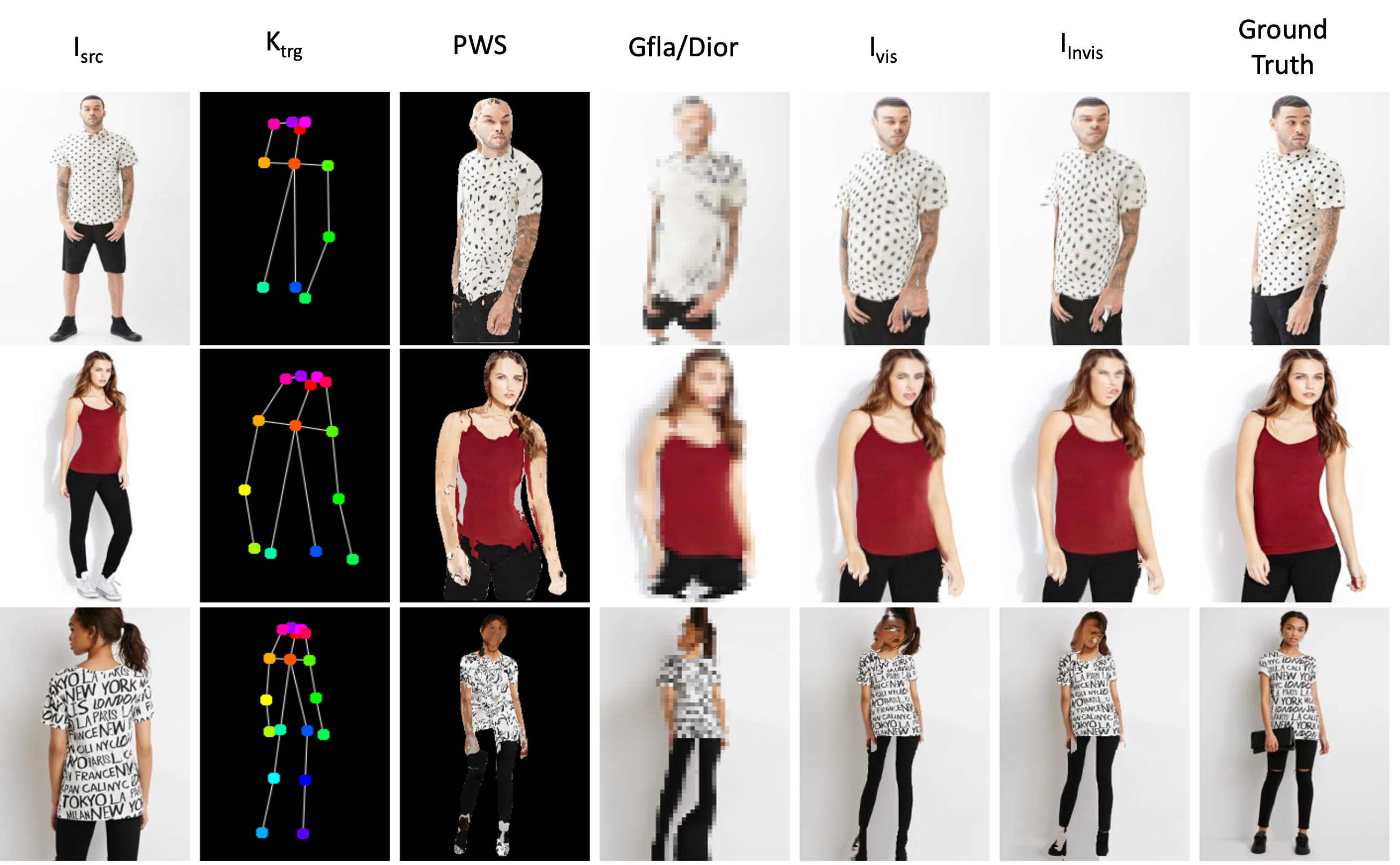}
\end{center}
\caption{Qualitative comparison between different warping functions for PWS\cite{albahar2021pose}, Gfla/Dior\cite{gfla, dior} and our ($I_{vis}$, $I_{Invis}$) warp. Gfla and Dior used the same flow prediction module} 
\label{fig:flowComp}
\end{figure}

\paragraph{Warping functions} Flow estimation is integral to many reposing pipelines\cite{gfla, dior, albahar2021pose}, including ours. Therefore we perform an analysis of the warping capabilities of different flow modules. The images in Fig \ref{fig:flowComp} show the quality of warping of the source image based on the flow predicted by the respective methods. As flow warping moves pixels, it can only hallucinate within the bound of the source image's content. We see that \OURNAME performs significantly better than previous flow-based warping techniques in preserving the semantic layout of the source human. The fine-grained texture details are transferred seamlessly along with the overall body shape. PWS is limited due to mistakes in estimating UV maps by off-the-shelf components\cite{densepose}, and GFLA\cite{gfla} had a sub-optimal flow estimation module. By letting our network predict different flow fields for visible and invisible regions while simultaneously constraining it to use the same latent features, we were able to achieve a good consistency between $I_{v}$ and $I_{i}$.

\begin{figure}[h!]
\begin{center}
  \includegraphics[width=\linewidth]{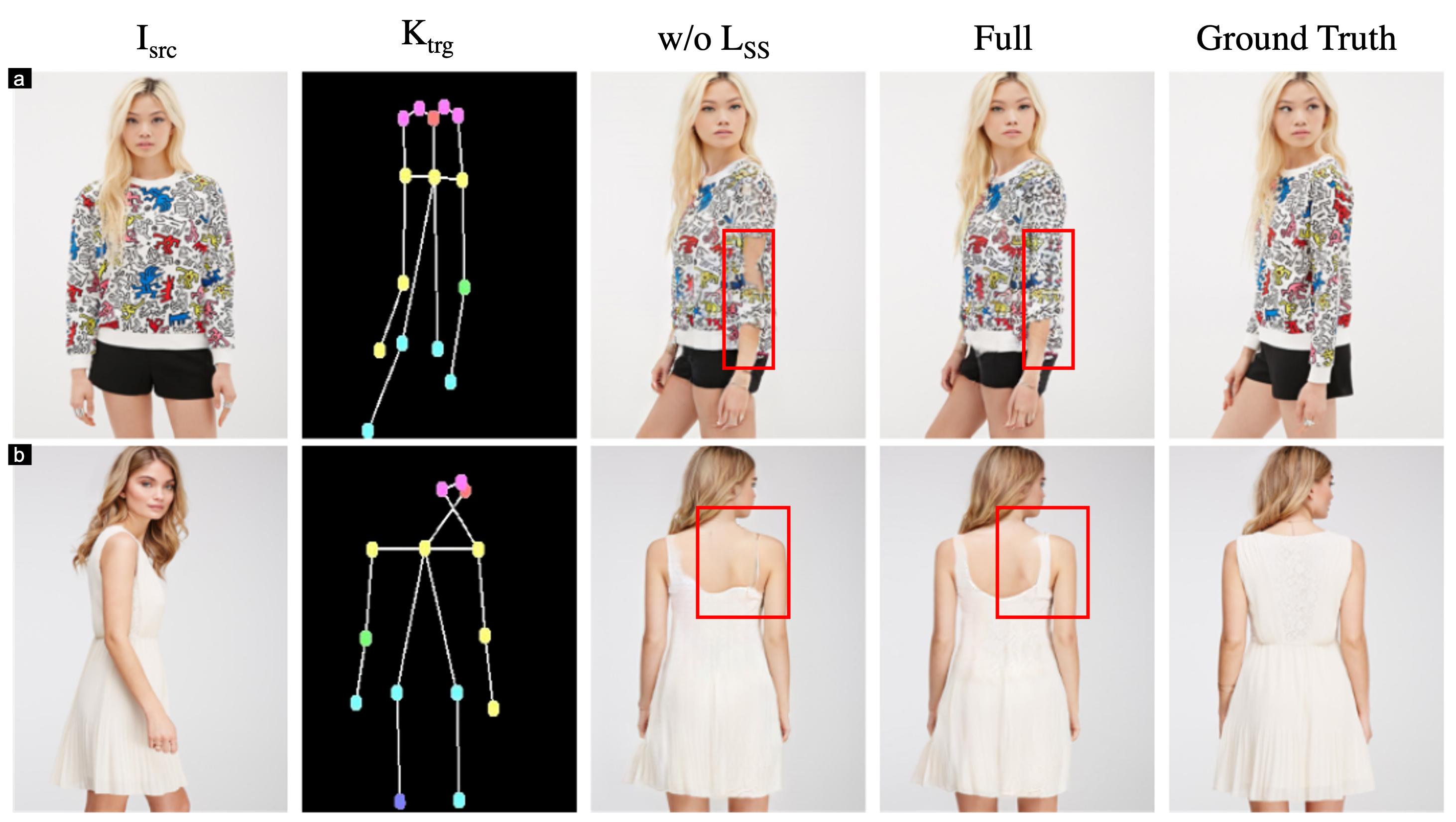}
\end{center}
\caption{Introducing PatchWise Self-Supervised Loss alleviates unnatural segmentation (a) and increases \textit{realness} (b)} 
\label{fig:ablation}
\end{figure}

\vspace{-4mm}
\paragraph{Ablation study} To gauge the effectiveness of different components of our pipeline, we perform various ablations of our network in Table \ref{tab:ablation}. Note that the Self Supervised loss is excluded in all the ablations. The result of ablations are as follow:-
\begin{itemize}[nosep]
    \item \textbf{w/o VisMap, $\boldsymbol{I_i,\  L_{SS}}$} The result of removing the Vismap and $I_i$ from the inputs of texture encoder and only passing $I_v$ indicates that the visibility map plays an integral role in capturing the appropriate relationship for the texture encodings.
    \item \textbf{w/o $\boldsymbol{I_i,\ L_{SS}}$} The degradation of quantitative image metrics on the removal of only the $I_i$ from the input of texture encoder shows that even though $I_v$ and $I_i$ produce similarly warped images, they do provide crucial complementary information.
    \item \textbf{w/o $\boldsymbol{k_s,\ L_{SS}}$} The deterioration in FID score(9.70 $\to$ 9.90) on removing the $K_s$ input from the pose encoding indicates that passing in $K_s$ helps in modeling the correlation between source and target pose.
    \item \textbf{w/o $\boldsymbol{L_{SS}}$} We also study the effect of finetuning with self-supervised loss $L_{SS}$. $L_{SS}$ plays a major role in improving the FID(9.70$\to$ 9.29) and marginally improving SSIM(0.725$\to$0.726) and LPIPS(0.186$\to$0.185). We also show qualitative improvements of integrating $L_{SS}$ in Fig \ref{fig:ablation}
    \item \textbf{Full} This model contains all the components presented in the paper. These ablations show that our configuration of \OURNAME produces the best output.
\end{itemize}

\begin{table}[t!]
\begin{center}
\begin{tabular}{|c|c|c|c|}
\hline
Method    & SSIM $\uparrow$  & FID $\downarrow$  & LPIPS $\downarrow$  \\
\hline
w/o VisMap, $I_{i}, L_{SS}$ & 0.719 & 9.89 & 0.196 \\
w/o $I_{i}, L_{SS}$ & 0.724 & 9.93 & 0.190 \\
w/o $K_{s}, L_{SS}$  & 0.726 & 9.90 & 0.186     \\
w/o $L_{SS}$  & 0.725 & 9.70 & 0.186  \\
Full & \textbf{0.726} & \textbf{9.29} & \textbf{0.185} \\
\hline
\end{tabular}
\end{center}
\caption{We perform extensive ablations to gauge the importance of each component in our network}
\label{tab:ablation}
\end{table}
\begin{figure}[h!]
\begin{center}
  \includegraphics[width=0.95\linewidth]{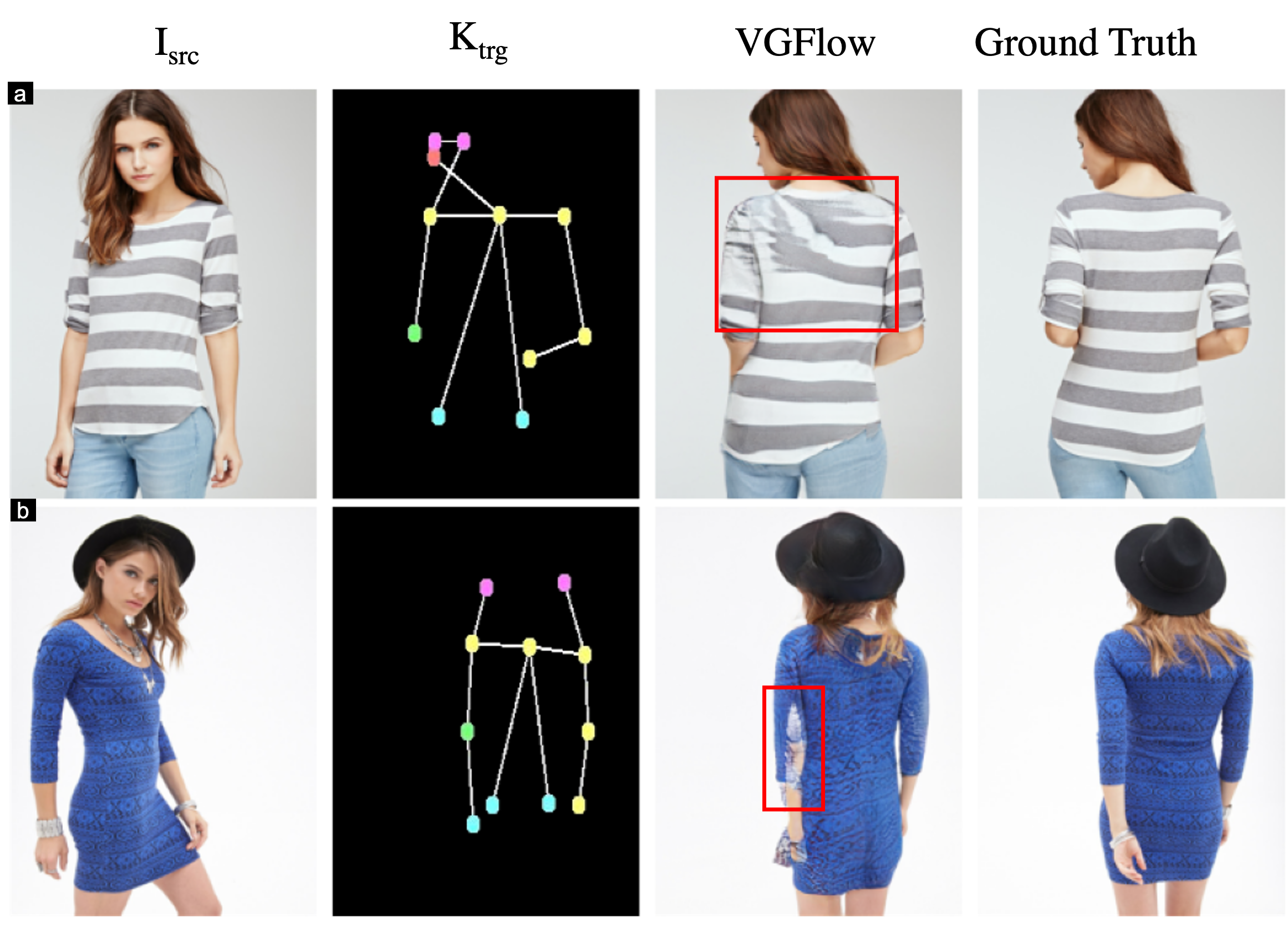}
\end{center}
\caption{Failure cases of \OURNAME due to warping limitations(a) and incorrect target pose segmentation(b)} 
\label{fig:failures}
\end{figure}

\vspace{-4mm}
\paragraph{Failures \& Limitations} Even though our network is able to produce accurate reposed outputs in majority of the cases, there are still limitations. In Fig \ref{fig:failures}, we see network artifacts that were caused due to incorrect warping (a) and erroneous target segmentation in \OURNAME(b). Some of these difficulties can be alleviated in future by enforcing symmetry\cite{albahar2021pose} or by leveraging target segmentation mask\cite{spgnet, dior}. 

\begin{figure}
\begin{center}
  \includegraphics[width=\linewidth]{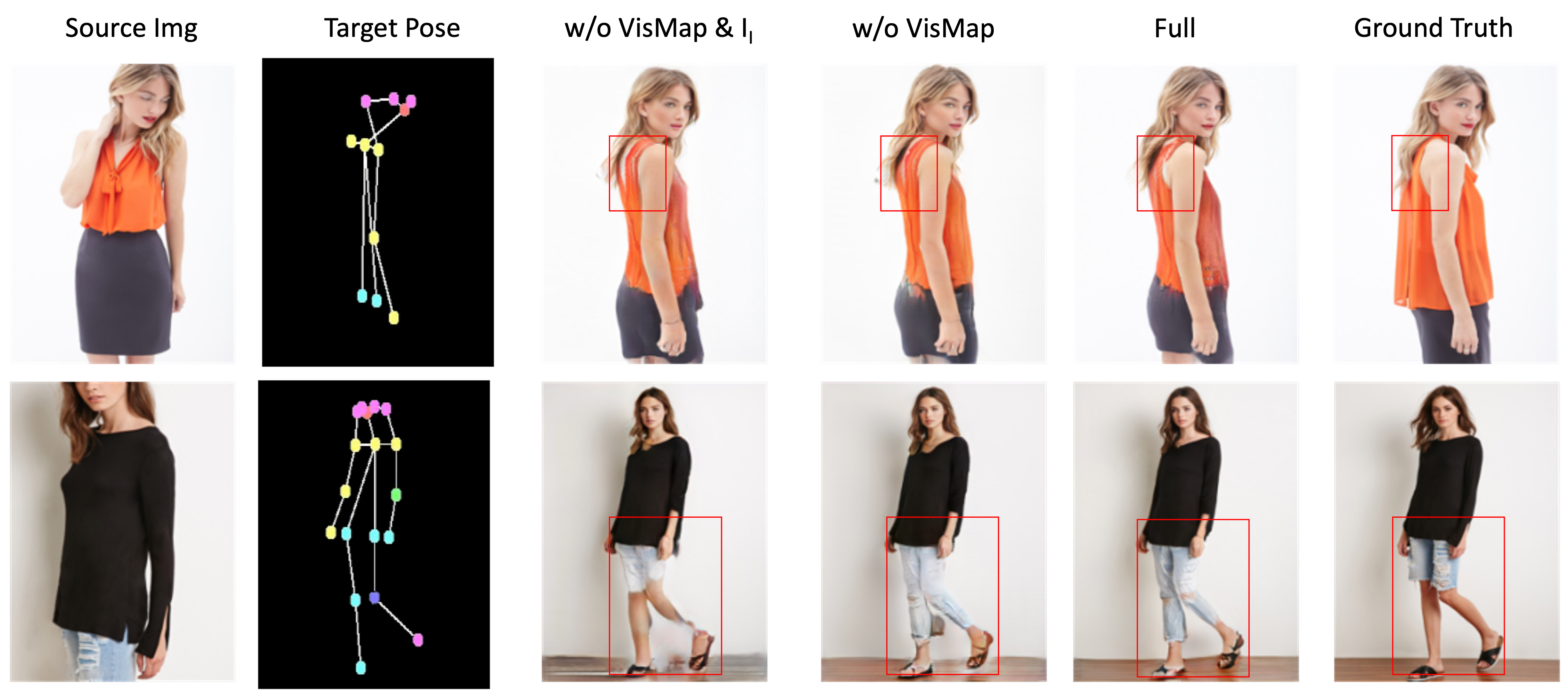}
\end{center}
\caption{Qualitative results for visibility map and flow fields ablation } 
\label{fig:visAblate}
\end{figure}

\vspace{-4mm}
\paragraph{Visibility ablation}
In Fig \ref{fig:visAblate}, we showcase the qualitative improvements achieved by VGFlow on passing Visibility map and both the warp outputs to the generator(quantitative improvement mentioned in paper ablation studies). The first column represents the source image, second column represent the target pose, third columns represents the results with only $I_{vis}$ input to the texture encoder. Fourth column was generated with $I_{vis}, VisMap$ as input and Full was generated with the complete pair of $I_{vis}, I_{Invis}, VisMap$ as input. In row 1, the network was able to complete the shirt design only after taking in all the inputs. Otherwise, it copied the front V-neck design to the back region. In row 2, the network produced artefacts while generating the lower portion of the garment which were fixed in the Full version.

\vspace{-4mm}
\paragraph{User Study} We conduct a user study with 25 volunteers to further support the performance capabilities of VGFlow. Participants  were shown 45 distinct result pairs sampled randomly from 8,570 test set results. Each pair consisted of one VGFlow result and the other sampled from the results of Dior/CASD baselines. These baselines were chosen to  represent keypoint based flow and attention based methods respectively. As can be seen from Tab \ref{tab:human-study}, VGFlow outperforms both the baselines in terms of preference by humans. We note that the majority of cases in which the users chose the baseline over VGFlow had minor segmentation issues in the invisible region, for which the humans are highly sensitive. These can be alleviated by incorporating the target segmentation mask in future.

\begin{table}
\begin{center}
\begin{tabular}{|c|c|c|}
\hline
Baseline & Prefer Baseline  &  Prefer VGFlow \\
\hline
\hline
Dior \cite{dior}   & 46\% & \textbf{54\%} \\
\hline
CASD \cite{casd}  & 37\%  & \textbf{63\%} \\
\hline
\end{tabular}
\end{center}
\caption{Survey results for gauging the human preference of \OURNAME over competing baselines. The percentage indicates the ratio of images which are voted to be better than the compared method.}
\label{tab:human-study}
\end{table}
\section{Conclusion}
We propose \OURNAME, a visibility-guided flow estimation network for human reposing that generates the reposed output by leveraging flow corresponding to different visibility regions of the human body. We propose an additional self-Supervised Patchwise GAN loss to reduce artifacts and improve the network's ability to adapt to various body shapes. \OURNAME achieves SOTA in the pose-guided person image generation task, and we demonstrate the significance of our contributions through extensive ablation studies.

{\small
\bibliographystyle{ieee_fullname}
\bibliography{egbib}
}

\pagebreak

\section{Additional results}

To highlight the superior quality output of VGFlow over previous baselines, we provide additional qualitative examples for the human reposing task in the subsequent pages.

\begin{figure*}
\begin{center}
  \includegraphics[width=\linewidth]{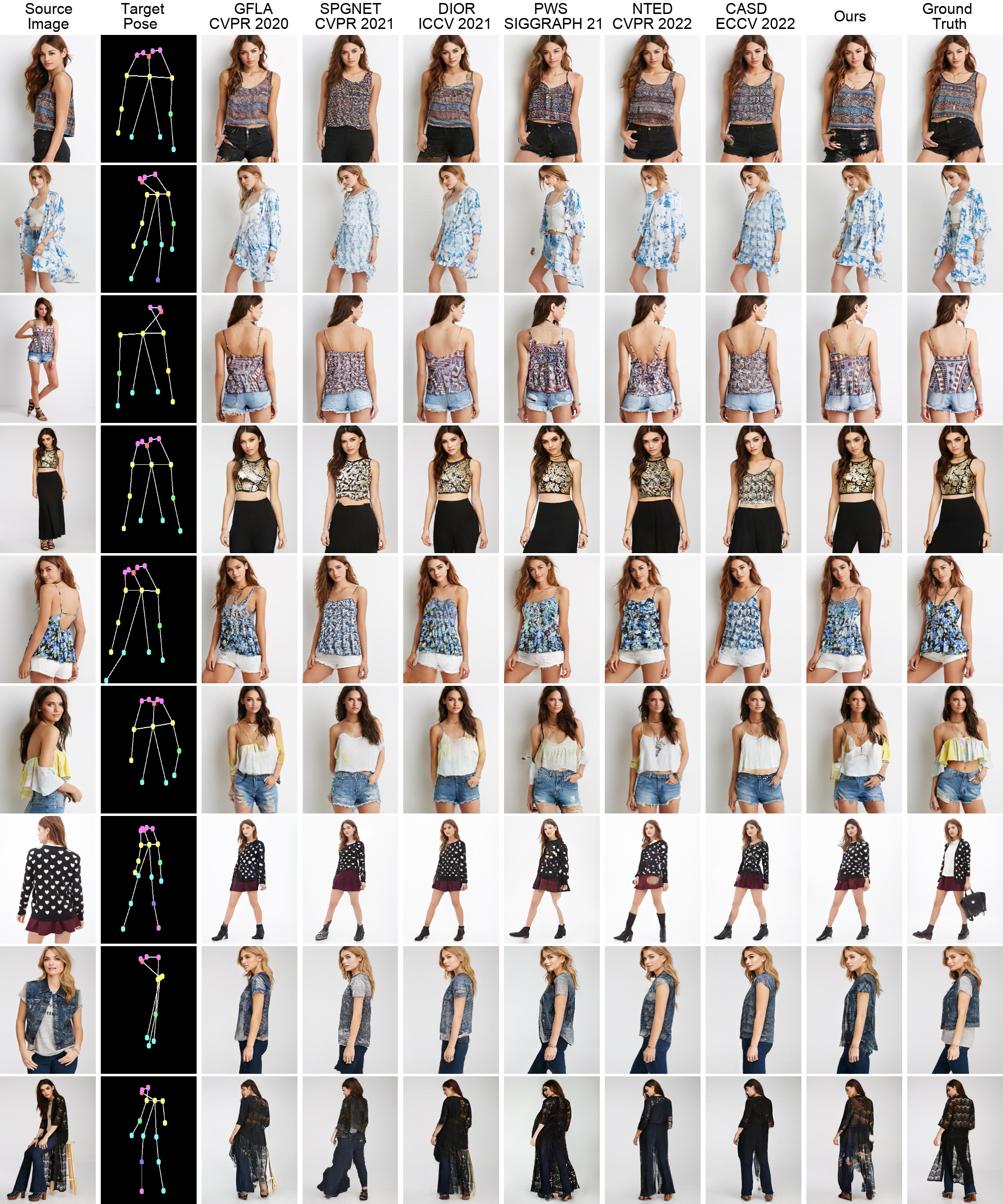}
\end{center}
\label{fig:qualitativeResults}
\caption{Qualitative results for human reposing}
\end{figure*}

\begin{figure*}
\begin{center}
  \includegraphics[width=\linewidth]{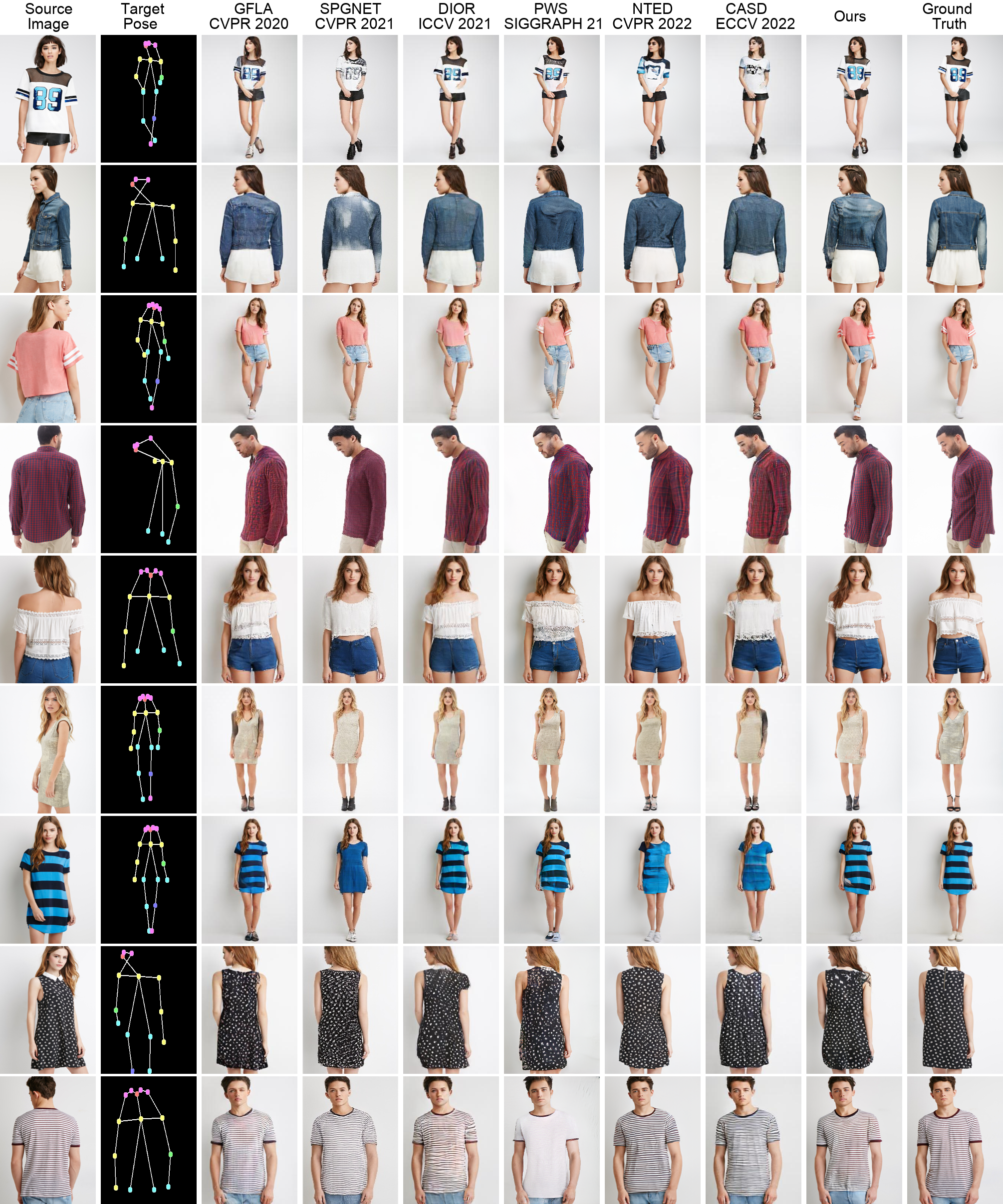}
\end{center}
\label{fig:qualitativeResults}
\caption{Qualitative results for human reposing}
\end{figure*}

\begin{figure*}
\begin{center}
  \includegraphics[width=\linewidth]{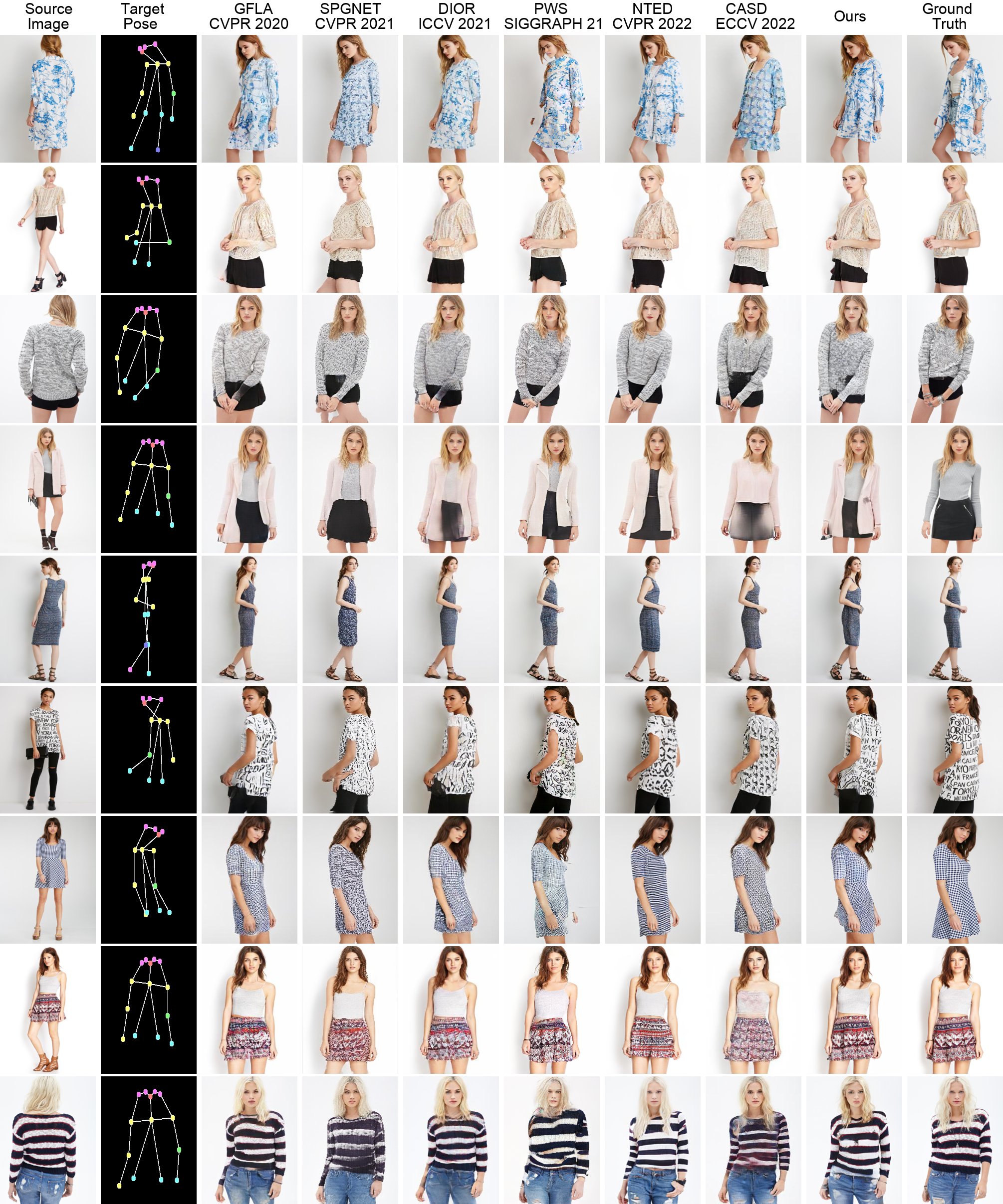}
\end{center}
\label{fig:qualitativeResults}
\caption{Qualitative results for human reposing}
\end{figure*}

\begin{figure*}
\begin{center}
  \includegraphics[width=\linewidth]{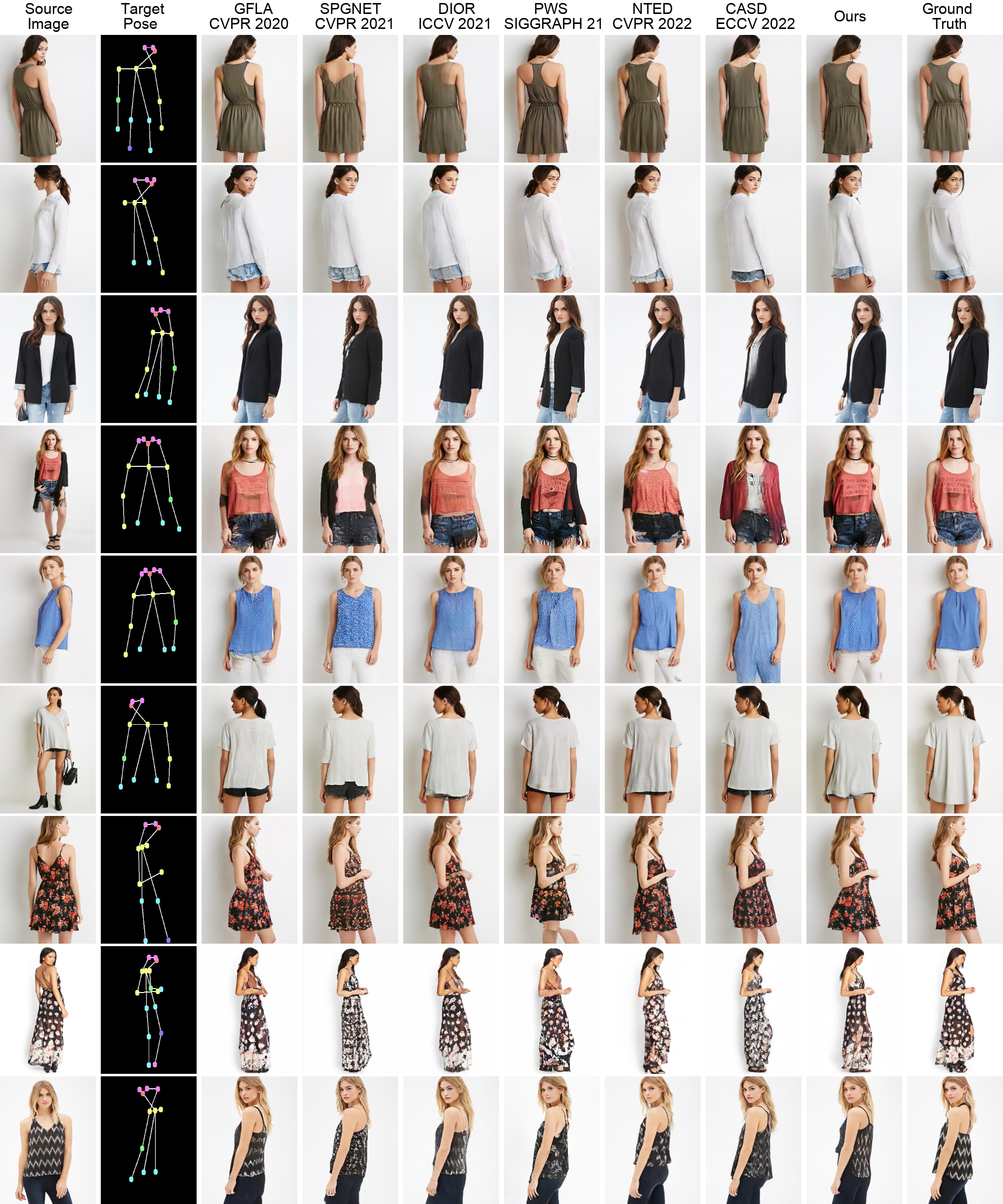}
\end{center}
\label{fig:qualitativeResults}
\caption{Qualitative results for human reposing}
\end{figure*}

\begin{figure*}
\begin{center}
  \includegraphics[width=\linewidth]{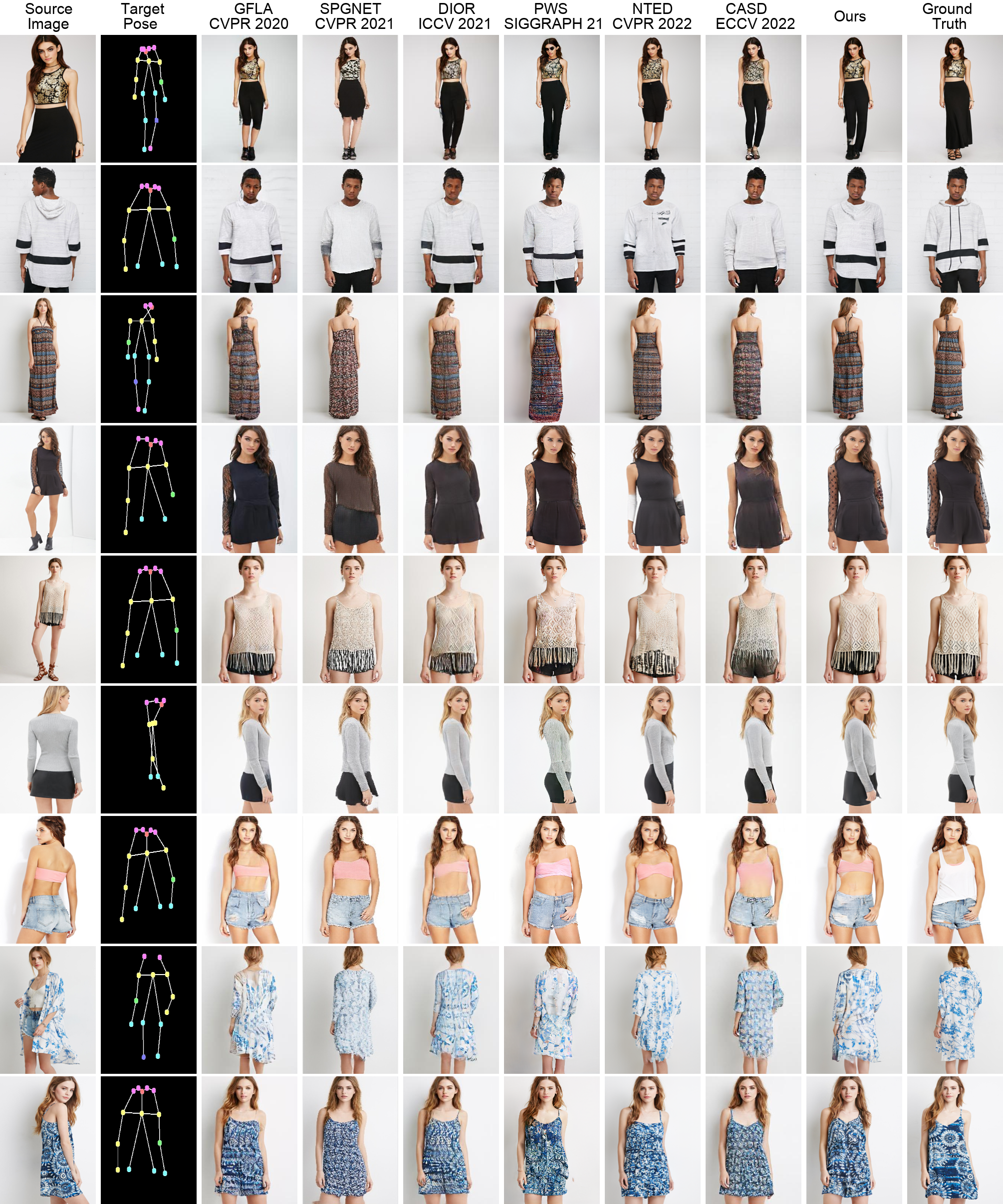}
\end{center}
\label{fig:qualitativeResults}
\caption{Qualitative results for human reposing}
\end{figure*}

\begin{figure*}
\begin{center}
  \includegraphics[width=\linewidth]{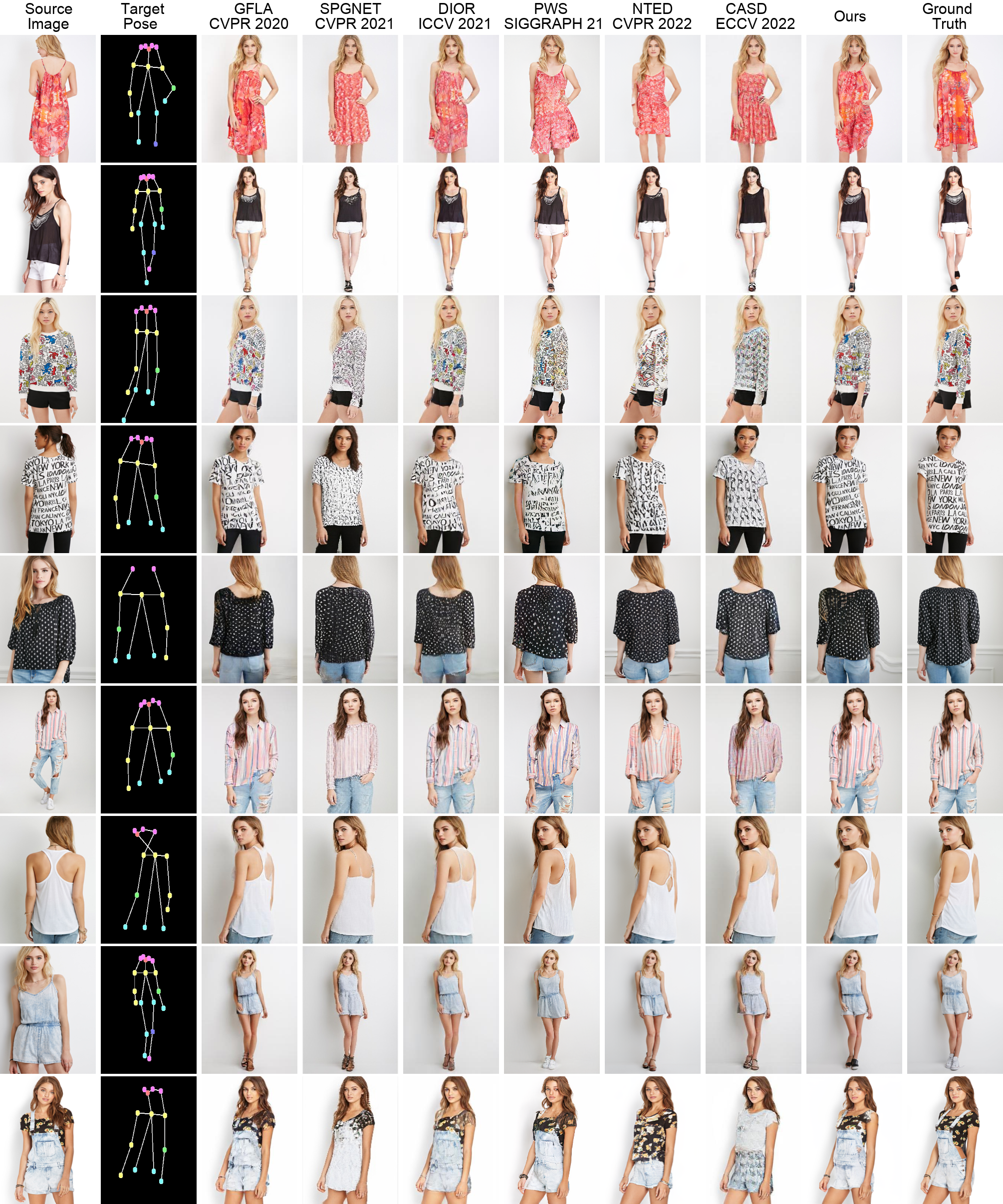}
\end{center}
\label{fig:qualitativeResults}
\caption{Qualitative results for human reposing}
\end{figure*}

\begin{figure*}
\begin{center}
  \includegraphics[width=\linewidth]{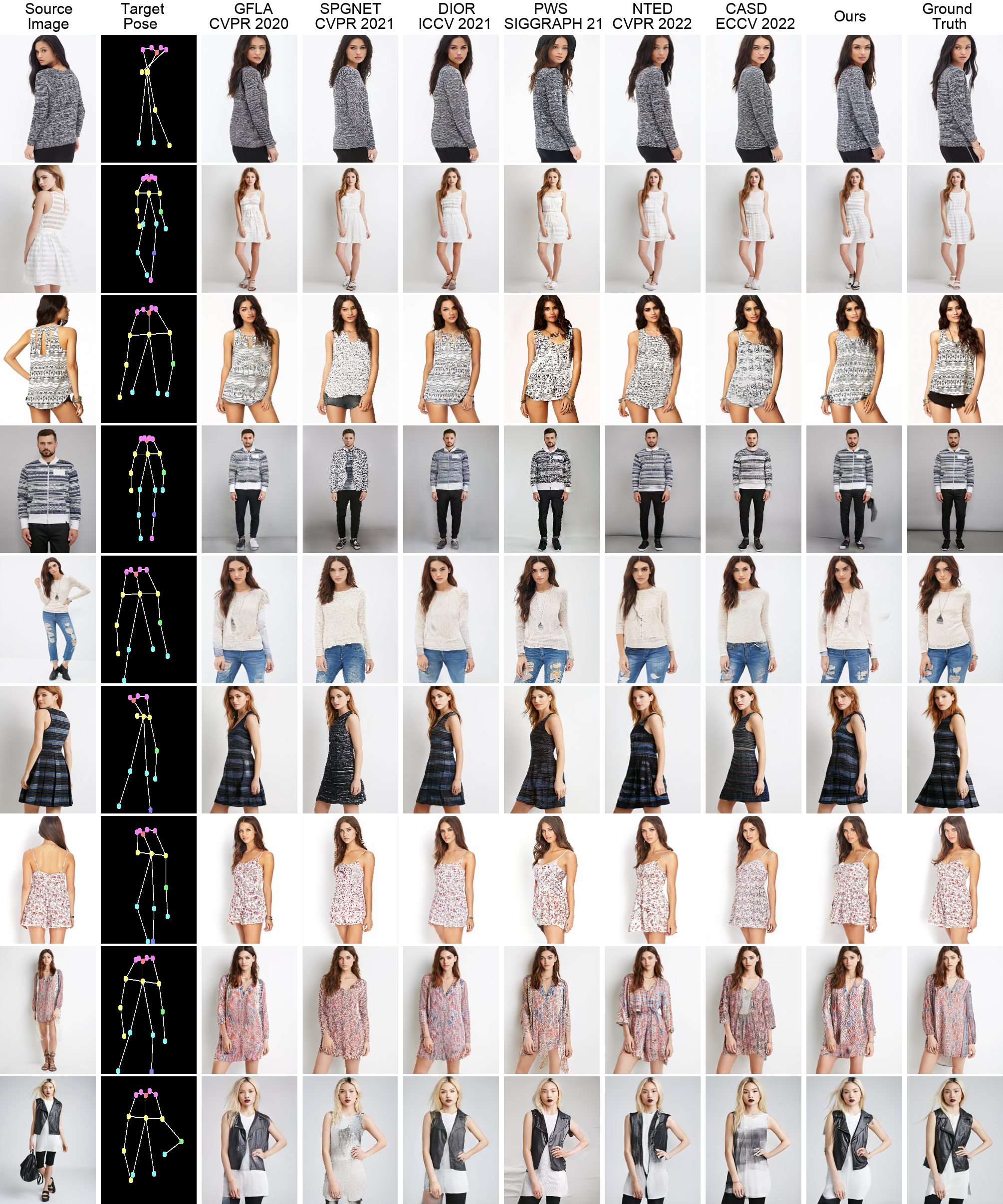}
\end{center}
\label{fig:qualitativeResults}
\caption{Qualitative results for human reposing}
\end{figure*}

\begin{figure*}
\begin{center}
  \includegraphics[width=\linewidth]{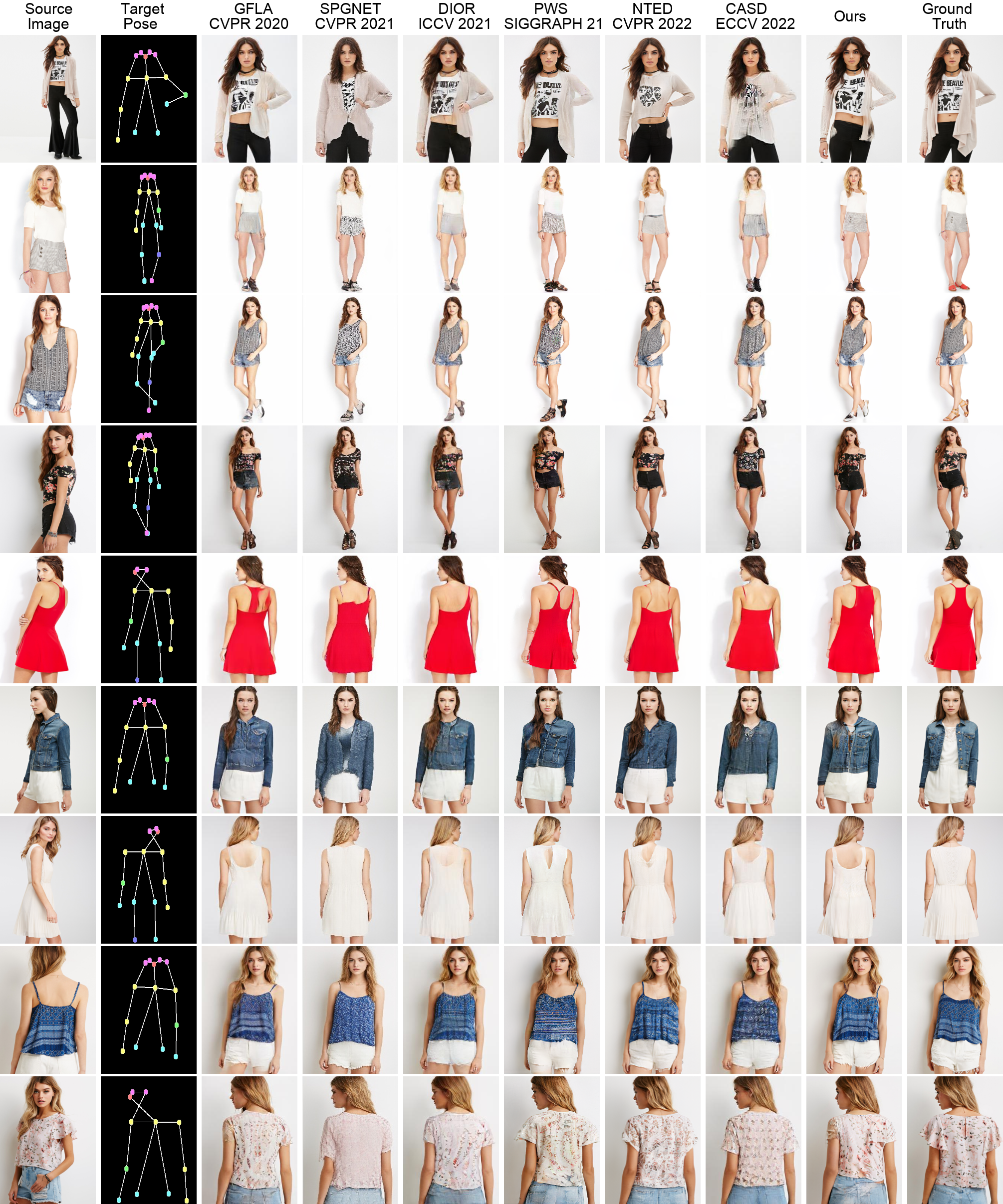}
\end{center}
\label{fig:qualitativeResults}
\caption{Qualitative results for human reposing}
\end{figure*}

\begin{figure*}
\begin{center}
  \includegraphics[width=\linewidth]{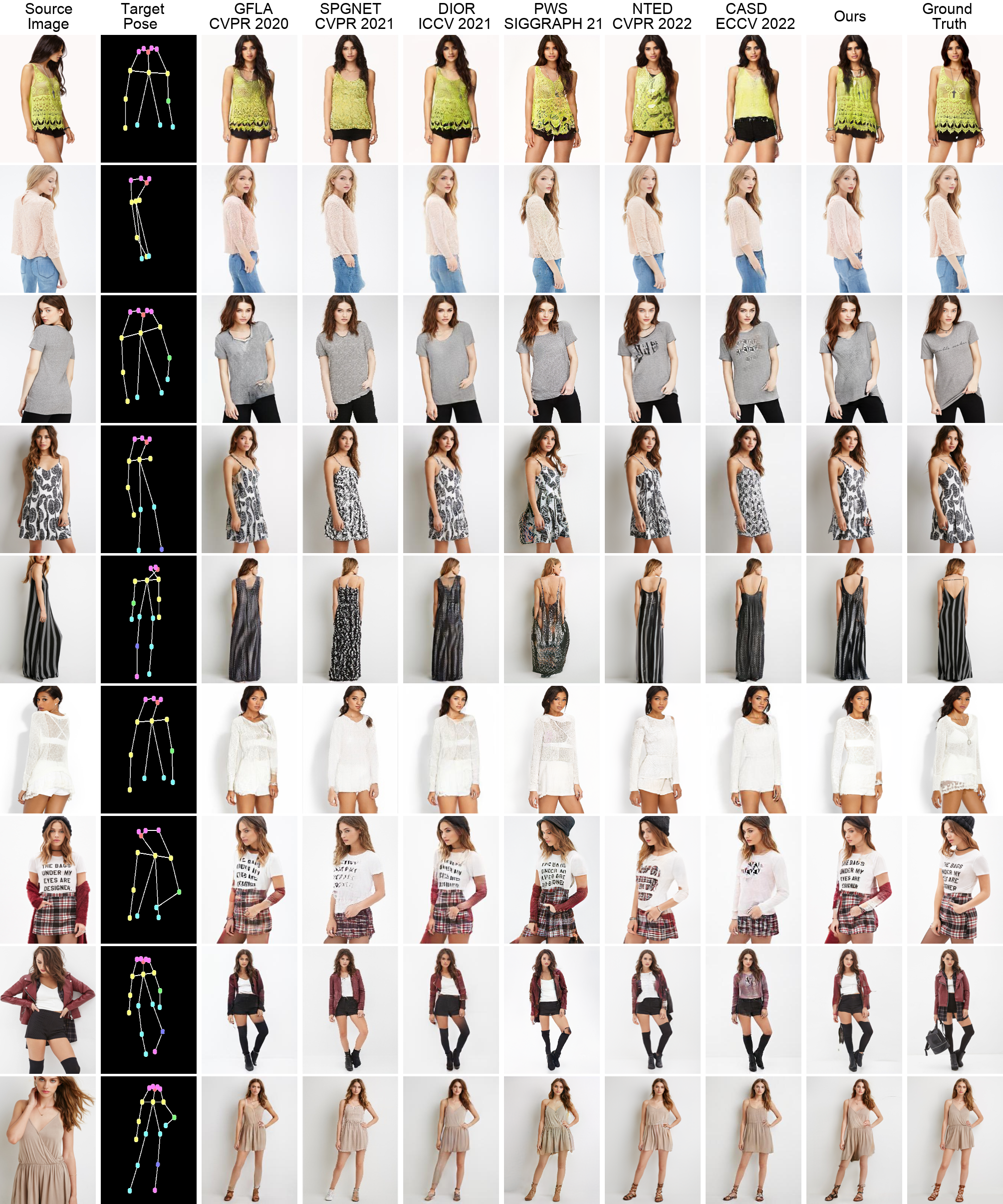}
\end{center}
\label{fig:qualitativeResults}
\caption{Qualitative results for human reposing}
\end{figure*}

\end{document}